\documentclass[10pt,twocolumn,letterpaper]{article}

\usepackage{graphicx}
\usepackage{amsmath}
\usepackage{amssymb}
\usepackage{booktabs}
\usepackage{subcaption}

\usepackage[pagebackref,breaklinks,colorlinks]{hyperref}

\usepackage[capitalize]{cleveref}
\crefname{section}{Sec.}{Secs.}
\Crefname{section}{Section}{Sections}
\Crefname{table}{Table}{Tables}
\crefname{table}{Tab.}{Tabs.}

\begin{document}

\title{MMPart: Harnessing Multi-Modal Large Language Models for Part-Aware 3D Generation}

\author{Omid Bonakdar\\
School of Computer\\
engineering, Iran university of\\
Science and Technology,\\
Tehran, Iran\\
{\tt\small omid\_bonakdar@comp.iust.ac.ir}
\and
Nasser Mozayani\\
School of Computer\\
engineering, Iran university of\\
Science and Technology,\\
Tehran, Iran\\
{\tt\small mozayani@iust.ac.ir}
}
\maketitle

\begin{abstract}
Generative 3D modeling has advanced rapidly, driven by applications in VR/AR, metaverse, and robotics. However, most methods represent the target object as a closed mesh devoid of any structural information, limiting editing, animation, and semantic understanding. Part-aware 3D generation addresses this problem by decomposing objects into meaningful components, but existing pipelines face challenges: in existing methods, the user has no control over which objects are separated and how model imagine the occluded parts in isolation phase. In this paper, we introduce MMPart, an innovative framework for generating part-aware 3D models from a single image. We first use a VLM to generate a set of prompts based on the input image and user descriptions. In the next step, a generative model generates isolated images of each object based on the initial image and the previous step’s prompts as supervisor (which control the pose and guide model how imagine previously occluded areas). Each of those images then enters the multi-view generation stage, where a number of consistent images from different views are generated. Finally, a reconstruction model converts each of these multi-view images into a 3D model.
\end{abstract}

\section{Introduction}
\label{sec:intro}

The demand for high-quality, interactive 3D content is accelerating at an unprecedented pace, driven by emerging technologies like virtual and augmented reality (VR/AR), the metaverse, digital twins, and advanced robotics. This has fueled a revolution in 3D generative modeling, where AI systems can now create impressive 3D assets from a variety of inputs, such as a line of text or a single 2D image \cite{xu2024instantmesh}\cite{liu2024meshformer}\cite{xu2023dmv3d}\cite{shi2023mvdream}\cite{tang2023dreamgaussian}\cite{siddiqui2024assetgen}.
However, a critical limitation persists in most state-of-the-art methods: they produce monolithic shapes. These assets are typically generated as a single, unstructured representation like a fused mesh or an implicit neural field without any distinct, meaningful components. This structural opaqueness presents a significant bottleneck for real-world applications. Tasks that are fundamental for artists and developers, such as compositional editing, procedural animation, material assignment, and semantic understanding, become difficult or impossible when an object cannot be broken down into its constituent parts.

\subsection{The Quest for Structure: An Overview of Part-Aware Generation}

To overcome the monolithic problem, the field is shifting its focus toward part-aware 3D generation. The goal is to create 3D objects not as indivisible blobs, but as compositions of semantically meaningful and geometrically complete parts, akin to how a human artist constructs a model. This approach unlocks the true potential of generative 3D, enabling assets that are inherently editable, animatable, and easier for downstream systems to understand

At the heart of this challenge lies a fundamental tension: the need to simultaneously achieve high structural cohesion, ensuring that all generated parts fit together to form a plausible and coherent whole, and low semantic coupling, where each part is distinct and can be addressed or manipulated independently. Navigating this balance has given rise to several distinct strategies for tackling the problem.

\subsection{Research Gap and Thesis Contribution}

The existing methodologies, while innovative, reveal a clear research gap. Multi-view pipelines\cite{part1232024}\cite{partgen2025} struggle with geometric consistency and occlusion, post-hoc completion methods\cite{holopart2025}\cite{omnipart2025} are sensitive to the quality of the initial 3D asset, and direct generation frameworks\cite{partcrafter2025} can be architecturally complex and data-hungry.
In this paper, we introduce an innovative framework for generating a part-aware 3D model from a single-view image, along with user descriptions (exactly what parts should be separated and how the occluded areas should be imagine), called MMPart. An input image with the given descriptions is first fed to a VLM, and the model combines its understanding of the input image, the user’s descriptive prompt, and our system prompt to generate separation guide prompts for the number of objects requested. These prompts describe the details of how each part should be isolated and how the occluded areas of each object should be imagined. In the next step, these prompts, along with the initial image, are fed to an image generator model to generate images of isolated objects requested by the user (In our similar methods, first multi-view images are generated and then a segmentation algorithm such as "SAM" is run on each image separately, which has challenges and shortcomings such as incoherent segmentations, then these regions enter the completion phase. Instead, in our method, separation is performed first and due to its generative nature, we do not need a completion phase, and then multi-view images are generated that are completely consistent.). Our method then uses the multiview propagation model to generate a set of multiview-compatible images and generates multiview-based images for each object. We then train a NURF model for each part on its corresponding multiview images. We then transform them into meshes using a mesh extraction algorithm such as flexicubes. partVlm leverages the power of VLMs in understanding and analyzing images as well as generating them in the process of generating part-aware 3D models, which makes it possible to compensate for many of the shortcomings of similar methods. For example, in similar methods, the user does not have control over exactly which objects are separated. Also, in the initial image, parts of an object may be occluded by another object, and these parts must be imagined in the isolation process. In other methods, the user does not have control over how these parts are imagined and must rely on the imagination of the complementary models themselves. However, in our method, the user can explain to the model exactly how he expects the occluded parts to be imagined.

\section{Related Works}
\label{sec:formatting}

\subsection{The Current Landscape: Strategies and Challenges}

Recent research has explored various pipelines for creating part-structured 3D assets, each with unique strengths and inherent limitations. These can be broadly categorized as follows:

Multi-View Segmentation and Reconstruction: A prominent strategy involves leveraging powerful, pre-trained 2D foundation models\cite{holopart2025}. These methods typically begin by generating multiple 2D views of an object\cite{part1232024}. Each view is then segmented into parts, and these 2D masks are "lifted" into 3D space during the final reconstruction phase.

Part123, for instance, generates multiview images and employs the Segment Anything Model (SAM) to get 2D part masks, using contrastive learning to manage inconsistencies during the 3D reconstruction process.

PartGen advances this idea by using one multi-view diffusion model to produce view-consistent part segmentations and a second model to perform amodal completion, filling in occluded regions of each part before reconstructing them in 3D13.

The primary challenge for these pipelines is overcoming the inconsistencies and ambiguities that arise from 2D views\cite{part1232024}. Errors in 2D segmentation can propagate and compound during 3D reconstruction, and accurately inferring the geometry of heavily occluded parts remains a significant hurdle\cite{omnipart2025}\cite{partgen2025}.

Post-hoc 3D Segmentation and Completion: An alternative approach begins with an already-generated monolithic 3D object. The model first segments the visible surface of the object into patches and then attempts to complete these patches into geometrically whole parts\cite{holopart2025}.

HoloPart pioneers this concept by introducing the task of 3D part amodal segmentation. It uses a diffusion-based model to intelligently "fill in the holes" of incomplete part segments, leveraging both local part geometry and global shape context.

While effective, this strategy's performance is heavily dependent on the quality of the initial 3D shape and the accuracy of the surface segmentation, which can be unreliable for meshes generated by other AI models\cite{omnipart2025}\cite{part1232024}.

Direct and Unified Compositional Generation: The latest wave of research focuses on generating all parts directly in 3D in an end-to-end or unified fashion, bypassing the need for explicit pre-segmentation.

OmniPart exemplifies a "plan-then-synthesize" strategy. It first employs an autoregressive model to plan a spatial layout of 3D bounding boxes for each part. Then, a spatially-conditioned generative model synthesizes all 3D parts simultaneously and consistently within this planned structure.

PartCrafter introduces a unified, one-shot architecture that takes a single image and jointly denoises multiple sets of latent tokens in parallel—one for each part. This allows for end-to-end, part-aware generation without relying on segmented inputs.

These direct approaches are powerful, as they avoid the error propagation issues of multi-view pipelines. However, they often require more complex architectural designs to manage the information flow between parts and depend on the availability of large-scale, part-annotated 3D datasets for training\cite{partcrafter2025}\cite{omnipart2025}.

\subsection{A Deeper Look at 2D-First Pipelines: Part123 and PartGen}

Among the strategies for part-aware generation, those based on multi-view 2D imagery have been particularly prominent due to their ability to leverage powerful, pre-existing 2D foundation models\cite{part1232024}\cite{partgen2025}.

Part123 and PartGen are two state-of-the-art works that exemplify this approach. While they have significantly advanced the field, a close examination of their methodologies reveals the inherent challenges of this 2D-first paradigm, directly motivating the work presented in this thesis.

Both frameworks follow a similar high-level pipeline: first, they generate multiple 2D images of an object from different viewpoints; second, they obtain 2D part segmentations for these images; and third, they use this 2D part information to guide a 3D reconstruction process.

Part123 tackles this by first using a multi-view diffusion model (syncDreamer\cite{liu2023syncdreamer}) to generate consistent images. It then employs the general-purpose Segment Anything Model (SAM)\cite{kirillov2023segmentanything} to produce segmentation masks for each view independently. The core challenge, as the authors note, is that these 2D masks are uncorrelated and inconsistent across views. For example, an object's head may be segmented as a single part in one view but grouped with the torso in another. To resolve this ambiguity, Part123 introduces contrastive learning into its neural rendering framework, training a feature field that learns to group pixels from the same mask together and separate those from different masks within the same view. The final 3D part segmentation is then derived by clustering these learned features.

PartGen\cite{partgen2025} addresses the segmentation problem more directly. Instead of relying on a general-purpose tool like SAM, it fine-tunes a multi-view diffusion model to perform stochastic, view-consistent part segmentation. This allows it to model the distribution of plausible part decompositions learned from artist-created data10. Its second key contribution is a contextual part completion module. Recognizing that parts are often occluded, it uses another diffusion model to complete the views of each part, taking the entire object as context to ensure the completed pieces fit together cohesively before reconstruction.
\subsubsection{Shortcomings and Challenges}
Despite their innovative solutions, both Part123 and PartGen are constrained by the fundamental limitations of their multi-stage, 2D-first pipeline:

Error Propagation: Both methods operate as a cascade of distinct steps. Any artifacts, inconsistencies, or failures in the initial multi-view generation will be passed down and amplified by the subsequent segmentation and reconstruction stages. Similarly, an inaccurate 2D segmentation can lead to a flawed final 3D model, regardless of the quality of the reconstruction network.

Ambiguity and Occlusion: Inferring 3D structure from 2D images is an ill-posed problem, especially for heavily occluded or entirely invisible parts. While PartGen's generative completion module is designed to "hallucinate" missing geometry, this remains a highly ambiguous task that is difficult to ground truthfully. Part123 must implicitly resolve these occlusions through its contrastive learning objective, which can be challenging when 2D mask information is sparse or contradictory.

Computational Complexity: These pipelines are computationally intensive. They require running multiple large-scale generative models in sequence—one for generating views, another for segmentation (and in PartGen's case, a third for completion)—followed by an optimization or feed-forward process for 3D reconstruction. This multi-model overhead can make the process slow and resource-heavy.

In summary, while Part123 and PartGen have demonstrated the potential of leveraging 2D priors, their reliance on a sequential, multi-stage pipeline creates inherent challenges related to error propagation, geometric ambiguity, and computational efficiency. This thesis aims to address these specific shortcomings by proposing a more integrated and robust framework.


\section{Methodology}

We introduce MMPart, a model for generating 3D objects with separate components by leveraging the power of multi modal llms.
Most existing multi-step methods use segmentation models such as SAM to separate different components, but we found that using multi modal llms in the component separation step can have many advantages and cover the shortcomings of previous methods.

\subsection{method architecture}

In this section, we discuss the architecture and working principle of MMPart. Our method receives an image along with a list of components desired by the user and generates a model with the separated components. The input image along with the user's desired component list is first entered into a VLM model. The VLM model analyzes the input image and, based on the component list, generates a prompt for each component, which is used in the next step. In this prompt, the image editing model is instructed to remove all components from the image except the desired object and provides a complete description of the pose, angle, lighting, and state of the isolated object. This accurate description ensures that the isolated object matches the original image perfectly. In the next step, a loop is executed for the desired number of objects, and each time the initial image is given to a generative model along with one of the prompts. This model produces an isolated image of that component, conditional on the initial image and according to the given prompt. In the next step, each of the generated isolated images is entered into a Multiview image generator, and new images are generated from them at different angles. These images are then used to build a 3D model of that component by a sparse view reconstruction model. Finally, we have the 3D model of each object along with its full texture, and we need to place them in the appropriate 3D position relative to each other based on the initial image.

\begin{figure*}
  \centering
  \begin{subfigure}{1\linewidth}
    \includegraphics[width=\linewidth]{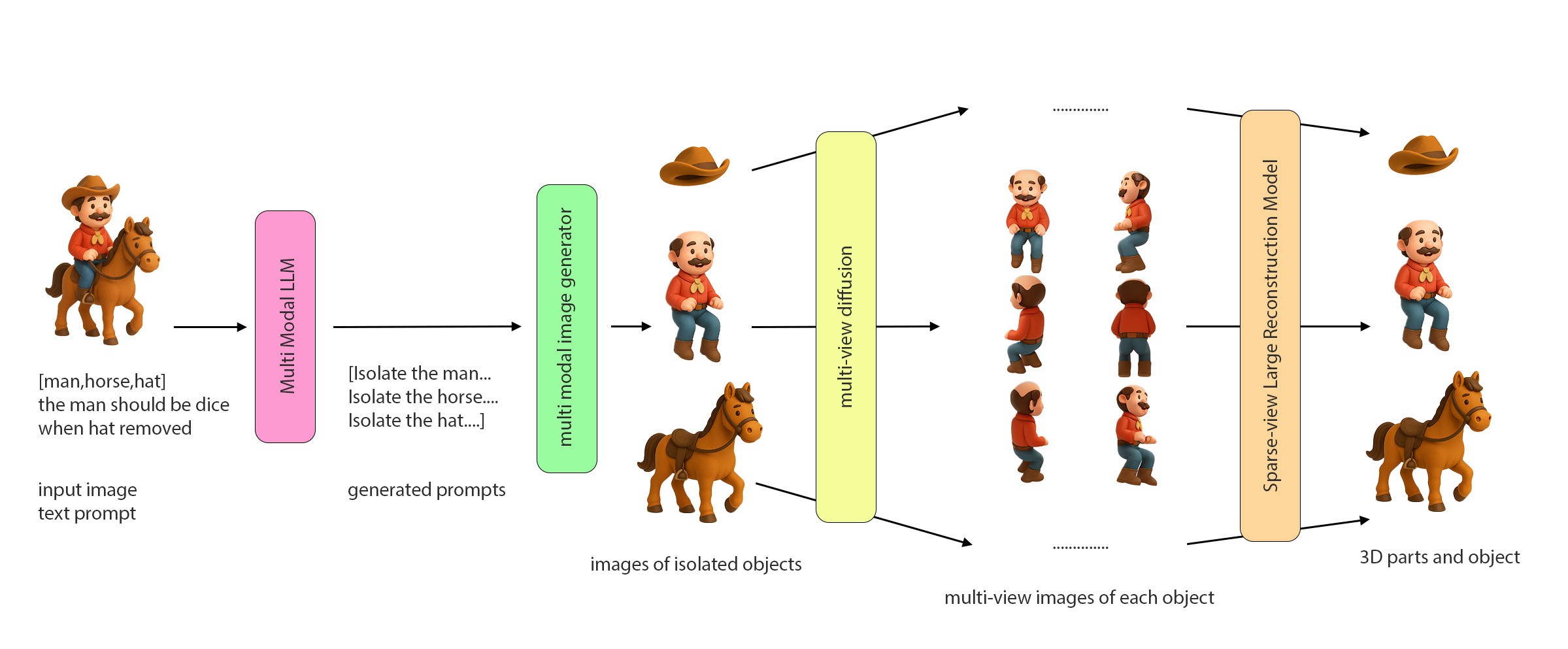} 
    \caption{Overall framework of our method}
    \label{fig:short-a}
  \end{subfigure}
  \hfill
\end{figure*}

\subsection{object and part isolation}
Similar works in this area use open world segmentation models such as SAM to separate components, but this method has major drawbacks, including that SAM segmentations are unlabeled, the user has no control over exactly which components are separated, different segmentations are created each time SAM is run, in the context of Multiview, segmentations in different views may be inconsistent, also some components may be completely or partially occluded, and to have complete objects, another step is needed to complete the incomplete images resulting from the segmentation. At this stage, without a proper prompt to guide the model, the model may be illusory and not produce a proper output. For example, if we segment a man sitting on a horse and then give it to an inpainting model to complete the incomplete parts created in the segmentation process, we expect it to deliver a man sitting in the air. However, because these models have rarely seen such images, we will see that most of the time, a chair or other object is placed under the man because it has been trained on such images. However, in our method, due to the presence of a prompt that describes the complete pose and state of the object, this does not happen and the model can imagine it well. Another point is that in the SAM method, the user has no control over how to complete the incomplete parts, and the completing model randomly imagines the covered parts each time, but in our method, the user has complete control over this issue. For example, imagine a man wearing a hat. The part of the man's head under the hat is the covered area that needs to be completed after separation. Now, different scenarios can be imagined. The man's head under the hat is bald, has little hair, or has a lot of hair. In the SAM method, the user has no control over this issue and can only achieve different results by running it multiple times. But in our method, the user can fully describe his request in the prompt stage. Right from the beginning, we can receive a guiding prompt from the user to describe exactly how these parts should be imagined. We found that new multi-modal image generation models such as image-1 are very powerful and can replace SAM-based approaches at this stage, while they do not have the drawbacks of the SAM method. The user can specify exactly which components they want to separate, which was not possible in SAM, and the only control the user had was the maximum number of regions. Because the individualization is done before the multiView generation step, the different views are fully compatible with each other. The Image-1 model, developed by OpenAI, is known as one of the advanced image generation models. This model is a multimodal system that is natively designed and, unlike many previous models that were only capable of processing text or generating images, it has the ability to receive text and images simultaneously as input. The main output of this model is the production or reproduction of high-quality images that are consistent with the input.
From an architectural perspective, Image-1 is based on the Transformer structure and processes text and image data in a common semantic representation space. In this process, the input text is decomposed into linguistic tokens and the input image is decomposed into visual tokens and then analyzed in a single architecture. Thus, the model is able to simultaneously understand the semantic relationships between text and visual data and include them in the output generation.
The main steps of image generation in the Image-1 model are as follows:
\begin{enumerate}
  \item Input processing: Converting text into linguistic tokens and image into visual tokens.
  \item Semantic fusion: Combining tokens in a common transformer architecture and creating a unified representation.
  \item Output generation: Generating a latent representation of the image and then reconstructing it into a high-resolution final image through reconstruction modules such as variational autoencoders (VAE) or diffusion-based models.
\end{enumerate}

Key capabilities of this model include:
\begin{itemize}
  \item Text-to-Image: Generating a new image based on text descriptions.
  \item Image Editing / Inpainting: Changing or recreating parts of an image based on textual input.
  \item Style Transfer: Reproducing an existing image in different artistic or visual styles.
  \item Text and Image Blending: Modifying or enhancing an existing image according to textual input.
\end{itemize}

The fundamental difference between Image-1 and models such as DALL·E 3 is that DALL·E was only able to receive text as input, while Image-1 was trained to be inherently multimodal. This feature allows the model to have a deeper and more accurate understanding of the combination of textual and visual inputs, and as a result, its outputs are much more optimized and natural in terms of visual quality, semantic coherence, and consistency with the user's command.

In addition to the Image-1 model, in recent years, numerous models with similar capabilities have been introduced, all of which fall into the category of multimodal generative models. Among these models, we can mention Imagen from Google, which is designed based on the Diffusion architecture and has the ability to produce high-quality images from text. Also, Stable Diffusion is provided as an open source model by Stability AI, and in addition to producing images from text, it also provides the ability to edit and reproduce images. Due to its open source nature, several optimized versions of it have been developed. MidJourney is also one of the most widely used systems in the field It is considered an image generation model that focuses more on artistic aspects and visual creativity. In addition, DeepFloyd IF is designed as a multi-stage (Cascade Diffusion) model and provides high accuracy in matching between text and image. The DALL·E 3 model (OpenAI product) is also considered the predecessor of Image-1, which had only text input and lacked a native multimodal structure. In addition, the FLUX.1 model (developed by Black Forest Labs in 2024) is also considered an important competitor that is based on a diffusion architecture and is designed to produce and edit high-quality images.
In general, although the aforementioned models have significant performance in the field of image generation or editing, the native multimodality feature of the Image-1 model makes this model stand out from many competitors in terms of architectural flexibility, output quality, and text-image matching.

\subsection{Multiview Diffusion}
Multiview diffusion models produce a set of consistent images from multiple angles using a single input image of an object for the purpose of 3D reconstruction.
This work employs \textbf{Zero123++}~\cite{shi2023zero123pp} to generate \textbf{$N=6$ images} from predetermined viewpoints arranged in a $3 \times 2$ tiling layout.
The underlying framework is based on diffusion models~\cite{Ho2020, SohlDickstein2015}, which treat the data distribution through a Markov chain of latent variables $z_0, z_1, \dots, z_T$.

The forward diffusion process incrementally introduces Gaussian noise to the initial data $z_0$ until only noise remains. This is defined by:
\begin{equation}
    q(z_{1:T}\mid z_0) = \prod_{t=1}^T \mathcal{N}\!\bigl(z_t; \sqrt{\alpha_t}z_{t-1}, (1 - \alpha_t)\mathbf{I}\bigr),
\end{equation}
where $\alpha_t$ governs the noise schedule. Conversely, the reverse process reconstructs data by denoising, beginning from a sample $z_T \sim \mathcal{N}(0,\mathbf{I})$. The joint distribution is:
\begin{equation}
    p(z_{0:T}) = p(z_T) \prod_{t=1}^T p_\theta(z_{t-1}\mid z_t),
\end{equation}
with
\begin{equation}
    p_\theta(z_{t-1}\mid z_t) = \mathcal{N}\!\bigl(z_{t-1}; \mu_\theta(z_t,t), \sigma_t^2\mathbf{I}\bigr).
\end{equation}
Here, a network learns the mean $\mu_\theta(z_t,t)$, while $\sigma_t$ is a constant derived from $\alpha_t$.

To ensure consistency across generated views, multiview diffusion extends this model to the joint distribution of all $N$ views. While the forward process adds noise to each view independently, the reverse process is generalized to denoise them collectively:
\begin{equation}
    p\!\bigl(z^{(1:N)}_{0:T}\bigr) = p\!\bigl(z^{(1:N)}_T\bigr) \prod_{t=1}^T p_\theta\!\bigl(z^{(1:N)}_{t-1}\mid z^{(1:N)}_t\bigr).
\end{equation}

\textbf{Zero123++}~\cite{shi2023zero123pp} implements this by tiling six fixed camera views into a single frame and introducing improved conditioning techniques—scaled reference attention for local conditioning and FlexDiffuse-style global conditioning—to leverage pretrained Stable Diffusion priors, resulting in high-quality, consistent multi-view generation from a single input image without explicit 3D supervision.

{\small
\bibliographystyle{ieee_fullname}
\bibliography{egbib}
}

\end{document}